\documentclass[a4paper, 10pt, conference]{ieeeconf}      

\IEEEoverridecommandlockouts                              
                                                          
\usepackage{mathptmx} 
\usepackage{times} 
\usepackage{amsmath} 
\usepackage{amssymb}  

\title{\LARGE \bf
Synthesis of Shared Control Protocols\\ with Provable Safety and
Performance Guarantees
}

\author{Nils Jansen$^{1}$ \and Murat Cubuktepe$^{1}$ \and Ufuk Topcu$^{1}$
\thanks{All authors are with the University of Texas at Austin, Austin, TX 78751, USA,
        {\tt\small njansen@utexas.edu}}%
}

\usepackage{amsmath,amssymb,amsfonts}
\usepackage{xspace}
\usepackage{todonotes}
\usepackage{colonequals}
\usepackage{booktabs}

\usepackage{subfigure}
\usepackage{url}
\usepackage{graphicx}
\usepackage{tikz} 
\usepackage{pgfplots}
\usepgfplotslibrary{external}
\usetikzlibrary{arrows,decorations.pathmorphing,positioning,fit,trees,shapes,shadows,automata,calc} 

\usepackage{macros} 

\tikzset{outline/.style args={#1}{%
  draw=#1,thick,fill=#1!50}}

\begin{document}

\maketitle
\thispagestyle{empty}
\pagestyle{empty}

\begin{abstract}
We formalize synthesis of shared control protocols with correctness guarantees for temporal logic specifications.
More specifically, we introduce a modeling formalism in which both a human and an autonomy
protocol can issue commands to a robot towards performing a certain
task. These commands are blended into a joint input to the robot. 
The autonomy protocol is synthesized using an abstraction of possible human commands accounting for randomness in decisions caused by factors such as fatigue or incomprehensibility of the problem at hand.
The synthesis is designed to ensure that the resulting robot behavior satisfies given safety and performance specifications,
e.g., in temporal logic. 
Our solution is based on nonlinear
programming and we address the inherent scalability issue by presenting
alternative methods. 
We assess the feasibility and the scalability of the approach by an experimental evaluation.
\end{abstract}

\section{Introduction}\label{sec:intro}
We study the problem of shared control, where a robot shall accomplish a task according to a human operator's goals and given specifications addressing safety or performance. Such scenarios are for instance found in remotely operated semi-autonomous wheelchairs~\cite{wheelchair-demillan}.  In a nutshell, the human has a certain action in mind and issues a \emph{command}. Simultaneously, an \emph{autonomy protocol} provides---based on the available information---another command. These commands are \emph{blended}---also referred to as \emph{arbitrated}---and deployed to the robot. 

Earlier work discusses shared control from different perspectives~\cite{dragan-et-al-assistive-teleoperation,dragan-et-al-policy-blending,DBLP:journals/corr/Trautman15a,DBLP:journals/corr/Trautman15,iturrate-et-al-shared-control-eeg,DBLP:journals/tase/FuT16}, however, \emph{formal correctness} in the sense of ensuring \emph{safety} or optimizing \emph{performance} has not been considered. In particular, having the human as an integral factor in this scenario, correctness needs to be treated in an appropriate way as a human might not be able to comprehend factors of a system and---in the extremal case---can drive a system into inevitable failure. 

There are several things to discuss. First, a human might not be sure about which command to take, depending on the scenario or factors like fatigue or incomprehensibility of the problem. We account for uncertainties in human decisions by introducing \emph{randomness} to choices. 
Moreover, a means of actually interpreting a command is needed in form of a user interface, \eg, a brain-computer interface; the usually imperfect interpretation adds to the randomness. We call a formal interpretation of the human's commands the \emph{human strategy} (this concept will be explained later).

  As many formal system models are inherently stochastic, our natural formal model for robot actions inside an environment is a Markov decision process (MDP) where deterministic action choices induce probability distributions over system states. Randomness in the choice of actions, like in the human strategy, is directly carried over to these probabilities when resolving nondeterminism. For MDPs, quantitative properties like ``the probability to reach a bad state is lower than $0.01$'' or ``the cost of reaching a goal is below a given threshold'' can be formally \emph{verified}. If a set of such \emph{specifications} is satisfied for the human strategy and the MDP, the task can be carried out \emph{safely} and \emph{with good performance}. 
  
  Given that the human strategy induces certain critical actions with a high probability, one or more specifications might be refuted. In this case, the autonomy should provide an alternative strategy that---when blended with the human strategy---satisfies the specifications without discarding too much of the human's choices. As in~\cite{dragan-et-al-policy-blending}, the blending puts weight on either the human's or the autonomy protocol's choices depending on factors such as the confidence of the human or the level of information the autonomy protocol has at its disposal.

The question is now how such a human strategy can be obtained.
  It seems unrealistic that a human can comprehend an MDP modeling a realistic scenario in the first place; primarily due the possibly very large size of the state space. Moreover, a human might not be good at making sense of probabilities or cost of visiting certain states at all. We employ \emph{learning} techniques to collect data about typical human behavior. This can, for instance, be performed within a simulation environment. In our case study, we model a typical shared control scenario based on a wheelchair~\cite{wheelchair-demillan} where a human user and an autonomy protocol share the control responsibility. Having a human user solving a task, we compute strategies from the obtained data using \emph{inverse reinforcement learning}~\cite{ng2000algorithms,abbeel2004apprenticeship}. Thereby, we can give guarantees on how good the obtained strategy approximates the actual intends of the user. 

The design of the autonomy protocol is the main concern of this paper. We define the underlying problem as a \emph{nonlinear optimization problem} and propose a technique to address the consequent scalability issues by reducing the problem to a \emph{linear optimization problem}. After an autonomy protocol is synthesized, guarantees on safety and performance can be given assuming that the user behaves according to the human strategy obtained beforehand.
The main contribution is a formal framework for the problem of shared autonomy together with thorough discussions on formal verification, experiments, and current pitfalls. A summary of the approaches and an outline are given in Section~\ref{sec:overview}.

Shared control has attracted considerable attention recently. We only overview some recent approaches into context with our results. First, Dragan and Srinivasa discussed strategy blending for shared control in~\cite{dragan-et-al-policy-blending,dragan-et-al-assistive-teleoperation}. There, the focus was on the \emph{prediction of human goals}. Combining these approaches, \eg, by inferring formal safety or performance specifications by prediction of human goals, is an interesting direction for future work.
 Iturrate \emph{et al.} presented shared control using feedback based on electroencephalography (a method to record electrical activity of the brain)~\cite{iturrate-et-al-shared-control-eeg}, where a robot is partly controlled via error signals from a brain-computer interface.
In~\cite{DBLP:journals/corr/Trautman15}, Trautman proposes to treat shared control broadly as a random process where different components are modeled by their joint probability distributions. As in our approach, randomness naturally prevents strange effects of blending: Consider actions ``up'' and ``down'' to be blended with equally distributed weight without having means to actually evaluating these weights. Finally, in~\cite{DBLP:journals/tase/FuT16} a synthesis method switches authority between a human operator and the autonomy such that satisfaction of linear temporal logic constraints can be ensured.

\section{Shared control}\label{sec:overview}
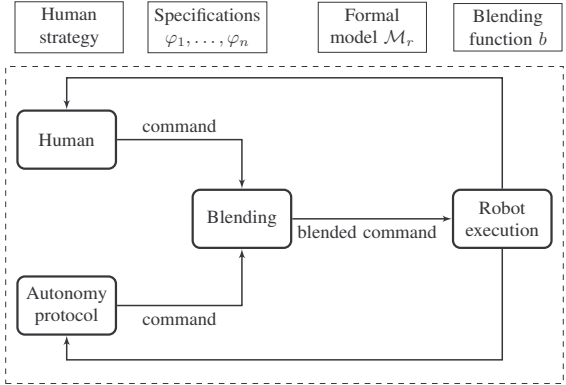
\begin{figure}[t]
\scalebox{0.7}{
	\centering
\begin{tikzpicture}
\tikzstyle{outer}= [draw, text centered, shape=rectangle, text width=1.8cm]
\tikzstyle{inner}=[draw, text centered, shape=rectangle, rounded corners, text width=1.5cm, minimum height=1.1cm, inner sep=5pt]			
\node[inner, very thick, text width=1.5cm] (human) {Human};


\node[inner, below=2cm of human, very thick, text width=1.5cm] (autonomy) {Autonomy\\ protocol};


\node[inner, right=1.4cm of human,yshift=-1.5cm, very thick, text width=1.5cm] (blending) {Blending};



\node[inner, right=3cm of blending, very thick, text width=1.5cm] (execution) {Robot\\ execution};


\draw[thick,-latex'] (human.east) -| node[above, near start] {command} (blending);


\draw[thick,-latex'] (autonomy.east)  -|  node[below, near start] (autonomyBlending) {command} (blending);



\draw (blending) edge[-latex', thick] node[below, text width=2.8cm] {blended command} (execution);







\draw[thick,-latex'] (execution.north) -- +(0,2.1) -| node [below, near start] {} coordinate (l1) (human.north);

\draw[thick,-latex'] (execution.south) -- +(0,-2.1) -| node [below, near start]  {} coordinate (l2) (autonomy.south);

%

\node (box) [draw, dashed, fit = (current bounding box), inner sep=0.2cm] {};

\node (input1) [outer,above=2.6cm of blending,xshift=5cm] {Blending function $\blendFunc$};
\node (input2) [outer,left=.5cm of input1] {Formal model $\mdp_r$};
\node (input3) [outer,left=3.5cm of input1, text width = 2 cm] {Specifications $\varphi_1,\ldots,\varphi_n$};
\node (input4) [outer,left=6.2cm of input1] {Human\\ strategy};


%

\end{tikzpicture}
}
\caption{Shared autonomy architecture.}
\label{fig:formal_setting}
\end{figure}
\noindent Consider first Fig.~\ref{fig:formal_setting} which recalls the general framework for shared autonomy with blending of commands; additionally we have a set of specifications, a formal model for robot behavior, and a blending function. 
In detail, a robot is to take care of a certain task. For instance, it shall move to a certain landmark. This task is subject to certain performance and safety considerations, \eg, it is not safe to take the shortest route because there are too many obstacles. These considerations are expressed by a set of \emph{specifications} $\varphi_1,\ldots,\varphi_n$. The possible behaviors of the robot inside an environment are given by a Markov decision process (MDP) $\mdp_r$. Having MDPs gives rise to choices of certain \emph{actions} to perform and to \emph{randomness} in the environment: A chosen path might induce a high probability to achieve the goal while with a low probability, the robot might slip and therefore fail to complete the task.

Now, in particular, a \emph{human user} issues a set of commands for the robot to perform. We assume that the commands issued by the human are consistent with an underlying randomized \emph{strategy $\sched_h$} for the MDP $\mdp_r$. Put differently, at design time we compute an abstract strategy $\sched_h$ of which the set of human commands is one realization. This modeling way
	allows to account for a variety of imperfections. Although it is not directly issued by a human, we call this strategy the \emph{human strategy}.
Due to possible human incomprehensibility or lack of detailed information, this leads to the fact that the strategy might not satisfy the requirements. 

Now, an \emph{autonomy protocol} is to be designed such that it provides an alternative strategy $\sched_a$, the \emph{autonomous strategy}. The two strategies are then \emph{blended}---according to the given blending function $\blendFunc$ into a new strategy $\sched_{ha}$ which satisfies the specifications. The blending function reflects preference over either the decisions of the human or the autonomy protocol. We also ensure that the blended strategy deviates only minimally from the human strategy. At runtime we can then blend decisions of the human user with decisions based on the autonomous strategy. The resulting ``blended'' decisions are according to the blended strategy $\sched_{ha}$, thereby ensuring satisfaction of the specifications. This procedure, while involving expensive computations at design time, is very efficient at runtime. 

Summarized, the problem we are addressing in this paper is then---in addition to the proposed modeling of the scenario---to \emph{synthesize} the autonomy protocol in a way such that the resulting blended strategy meets \emph{all of the specifications} while it only deviates from the human strategy as little as possible. 
	We introduce all formal foundations that we need in Section~\ref{sec:preliminaries}. The \emph{shared control synthesis problem} with all needed formalisms is presented in Section~\ref{sec:synthesisproblem} as being a \emph{nonlinear optimization problem}. Addressing scalability, we reduce the problem to a \emph{linear optimization problem} in Section~\ref{sec:solution}. We indicate the feasibility and scalability of our techniques using data-based experiments in Section~\ref{sec:solution} and draw a short conclusion in Section~\ref{sec:conclusion}.

\section{Preliminaries}\label{sec:preliminaries}
%
\subsubsection{Models}
A \emph{probability distribution} over a finite or countably infinite set $\distDom$ is a function $\distFunc\colon\distDom\rightarrow\Ireal$ with $\sum_{\distDomElem\in\distDom}\distFunc(\distDomElem)=\distFunc(\distDom)=1$. 
The set of all distributions on $\distDom$ is denoted by $\Distr(\distDom)$.

\begin{definition}[MDP]
A \emph{Markov decision process (MDP)} $\MdpInit$ is a tuple with a set of states $S$, a unique initial state $\sinit \in S$, a finite set $\Act$ of actions, and a (partial) probabilistic transition function $\probmdp\colon S\times\Act\rightarrow\Distr(S)$.
\end{definition}
MDPs operate by means of \emph{nondeterministic choices} of actions at each state, whose successors are then determined \emph{probabilistically} with respect to the associated probability distribution.
The \emph{enabled} actions at state $s\in S$ are denoted by $\Act(s)=\{\act\in\Act\mid\exists\mu\in\Distr(S).\,\mu=\probmdp(s,\act)\}$. 
To avoid deadlock states, we assume that $|\Act(s)|\geq 1$ for all $s\in S$.
A \emph{cost function} $\rho\colon S\times\Act\rightarrow\R_{\geq 0}$ for an MDP $\mdp$ adds cost to a \emph{transition} $(s,\act)\in S\times\Act$ with $\act\in\Act(s)$.  
A \emph{path} in an $\mdp$ is a finite (or infinite) sequence $\pi=s_0\act_0s_1\act_1\ldots$ with $\probmdp(s_i, \act, s_{i+1}) > 0$ for all $i\geq 0$. 
%
If $|\Act(s)|=1$ for all $s\in S$, all actions can be disregarded and the MDP $\mdp$ reduces to a \emph{discrete-time Markov chain (MC)}.

The \emph{unique probability measure} $\pr^\dtmc(\Pi)$ for a set $\Pi$ of paths of MC $\dtmc$ can be defined by the usual cylinder set construction, the \emph{expected cost} of a set $\Pi$ of paths is denoted by $\expRew^\dtmc(\Pi)$,
 see~\cite{BK08} for details.
In order to define a probability measure and expected cost on MDPs, the nondeterministic choices of actions are resolved by so-called \emph{strategies}. 
For practical reasons, we restrict ourselves to \emph{memoryless} strategies, again refer to~\cite{BK08} for details.
\begin{definition}[Strategy]\label{def:scheduler}
	A \emph{randomized strategy} for an MDP $\mdp$ is a function $\sched\colon S\rightarrow\Distr(\Act)$ such that $\sigma(s)(\act) > 0$ implies $\act \in \Act(s)$. A strategy with $\sched(s)(\act)=1$ for $\act\in\Act$ and $\sched(\beta)=0$ for all $\beta\in\Act\setminus\{\act\}$ is called \emph{deterministic}. The set of all strategies over $\mdp$ is denoted by $\Sched^\mdp$.
\end{definition}
%
Resolving all nondeterminism for an MDP $\mdp$ with a strategy $\sched\in\Sched^\mdp$ yields an \emph{induced Markov chain} $\mdp^\sched$. Intuitively, the random choices of actions from $\sched$ are transferred to the transition probabilities in $\mdp^\sched$.
\begin{definition}[Induced MC]\label{def:induced_dtmc}
	Let MDP $\MdpInit$ and strategy $\sched\in\Sched^{\mdp}$. The \emph{MC induced by $\mdp$ and $\sched$} is  $\mdp^\sched=(S,\sinit,\Act,\probmdp^\sched)$ where
	\begin{align*}
		\probmdp^\sched(s,s')=\sum_{\act\in\Act(s)} \sched(s)(\act)\cdot\probmdp(s,\act)(s') \mbox{ for all } s,s'\in S\ .
	\end{align*} 
\end{definition}
%
%
%
%
\subsubsection{Specifications}
A quantitative reachability property $\reachProplT$ with upper probability threshold $\lambda\in\Irat$ and target set $T\subseteq S$ constrains the probability to reach $T$ from $\sinit$ in $\mdp$ to be at most $\lambda$. 
Expected cost properties $\expRewProp{\kappa}{G}$ impose an upper bound $\kappa\in\Q$ on the expected cost to reach goal states $G\subseteq S$. 
Intuitively, bad states $T$ shall only be reached with probability $\lambda$ (\emph{safety specification}) while the expected cost for reaching goal states $G$ has to be below $\kappa$ (\emph{performance specification}). 
Probability and expected cost to reach $T$ from $\sinit$ are denoted by $\pr(\finally T)$ and $\expRew(\finally T)$, respectively. Hence, $\pr^{\dtmc}(\finally T)\leq\lambda$ and $\expRew^{\dtmc}(\finally G)\leq\kappa$ express that the properties $\reachProplT$ and $\expRewProp{\kappa}{G}$ are satisfied by MC $\dtmc$. These concepts are analogous for lower bounds on the probability. We also use \emph{until properties} of the form $\pr_{\geq\lambda} (\neg T\ \pctlUntil\ G)$ expressing that the probability of reaching $G$ while not reaching $T$ beforehand is at least $\lambda$.

An MDP $\mdp$ satisfies both safety specification $\phi$ and performance specification $\psi$, iff \emph{for all} strategies $\sched\in\Sched^\mdp$ it holds that the induced MC $\mdp^\sched$ satisfies $\phi$ and $\psi$, i.e., $\mdp^\sched\models\phi$ and $\mdp^\sched\models\psi$. If several performance or safety specifications $\varphi_1,\ldots,\varphi_n$ are given MDP $\mdp$, the simultaneous satisfaction for all strategies, denoted by $\mdp\models\varphi_1,\ldots,\varphi_n$, can be formally verified for an MDP using multi-objective model checking~\cite{DBLP:journals/lmcs/EtessamiKVY08}. 

Here, we are interested in the \emph{synthesis problem}, where the aim is to find one particular strategy $\sched$ for which the specifications are satisfied.
If for $\varphi_1,\ldots,\varphi_n$ and strategy $\sched$ it holds that $\mdp^\sched\models\varphi_1,\ldots,\varphi_n$, then $\sched$ is said to \emph{admit} the specifications, also denoted by $\sched\models\varphi_1,\ldots,\varphi_n$.
\begin{figure}[t]
	\centering
	\subfigure[MDP $\mdp$]
	{
		\scalebox{0.7}{\begin{tikzpicture}[scale=1, nodestyle/.style={draw,circle},baseline=(s0)]

    \node [nodestyle] (s0) at (0,0) {$s_0$};
    \node [nodestyle] (s1) [on grid, right=2.5cm of s0] {$s_1$};
    \node [nodestyle, accepting] (s2) [on grid, right=2.5cm of s1] {$s_2$};
    \node [nodestyle] (s3) [on grid, above=2.5cm of s1] {$s_3$};
    \node [nodestyle] (s4) [on grid, above=2.5cm of s2] {$s_4$};
    
    \node [circle, draw, scale=0.5, fill=black] (s0a) [right=.7cm of s0] {};
    \node [circle, draw, scale=0.5, fill=black] (s0b) [above=0.8cm of s0a] {};
    \node [circle, draw, scale=0.5, fill=black] (s1a) [right=.7cm of s1] {};
    \node [circle, draw, scale=0.5, fill=black] (s1b) [above=0.8cm of s1a] {};

    \draw [thick] (s0) -- node[above] {$a$} (s0a);
    \draw [thick] (s0) -- node[above] {$b$} (s0b);
    \draw [thick] (s1) -- node[above] {$c$} (s1a);
    \draw [thick] (s1) -- node[above] {$d$} (s1b);
    
    \draw[->] (s0a) -- node [auto] {\scriptsize$0.6$} (s1);
    \draw[->] (s0a) -- node [right, near end] {\scriptsize$0.4$} (s3);
    \draw[->] (s0b) -- node [auto] {\scriptsize$0.4$} (s1);
    \draw[->] (s0b) -- node [auto] {\scriptsize$0.6$} (s3);
    
    \draw[->] (s1a) -- node [auto] {\scriptsize$0.6$} (s2);
    \draw[->] (s1a) -- node [right, near end] {\scriptsize$0.4$} (s4);
    \draw[->] (s1b) -- node [auto] {\scriptsize$0.4$} (s2);
    \draw[->] (s1b) -- node [auto] {\scriptsize$0.6$} (s4);
    
    \draw(s4) edge[loop above] node [above] {\scriptsize$1$} (s4);
    \draw(s3) edge[loop above] node [above] {\scriptsize$1$} (s3);
    \draw(s2) edge[loop above] node [above] {\scriptsize$1$} (s2);    
    
   \end{tikzpicture}}\label{fig:mdp}
	}
	\subfigure[Induced MC $\mdp^{\sched_1}$]
	{
		\scalebox{0.7}{\begin{tikzpicture}[scale=1, nodestyle/.style={draw,circle},baseline=(s0)]

    \node [nodestyle] (s0) at (0,0) {$s_0$};
    \node [nodestyle] (s1) [on grid, right=2cm of s0] {$s_1$};
    \node [nodestyle, accepting] (s2) [on grid, right=2cm of s1] {$s_2$};
    \node [nodestyle] (s3) [on grid, above=2.5cm of s1] {$s_3$};
    \node [nodestyle] (s4) [on grid, above=2.5cm of s2] {$s_4$};
    

    
    \draw[->] (s0) -- node [auto] {\scriptsize$0.6$} (s1);
    \draw[->] (s0) -- node [right, near end] {\scriptsize$0.4$} (s3);
    
    \draw[->] (s1) -- node [auto] {\scriptsize$0.6$} (s2);
    \draw[->] (s1) -- node [right, near end] {\scriptsize$0.4$} (s4);
    
    \draw(s4) edge[loop above] node [above] {\scriptsize$1$} (s4);
    \draw(s3) edge[loop above] node [above] {\scriptsize$1$} (s3);
    \draw(s2) edge[loop above] node [above] {\scriptsize$1$} (s2);    
    
   \end{tikzpicture}}\label{fig:induced_dtmc1}
	}
	\caption{MDP $\mdp$ with target state $s_2$ and induced MC for strategy $\sched_{\textit{unif}}$}
	\label{fig:mdp_example}
\end{figure}
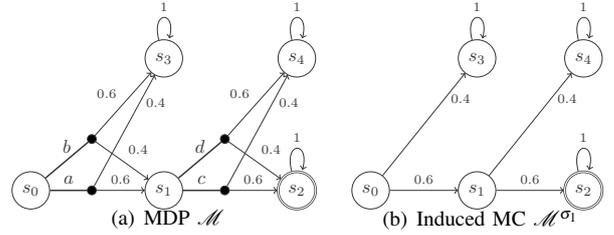
\begin{example}\label{ex:simple_mdp}
Consider Fig.~\ref{fig:mdp} depcting MDP $\mdp$ with initial state $s_0$, where states $s_0$ and $s_1$ have choices between actions $a$ or $b$ and $c$ or $d$, respectively. For instance, action $a$ induces a probabilistic choice between $s_1$ and $s_3$ with probabilities $0.6$ and $0.4$. The self loops at $s_2,s_3$ and $s_4$ indicate looping back with probability one for each action.

Assume now, a safety specification is given by $\phi=\reachProp{0.21}{s_2}$. The specification is \emph{violated} for $\mdp$, as the deterministic strategy $\sched_{1}\in\Sched^\mdp$ with $\sched_{1}(s_1)(\act)=1$ and $\sched_{1}(s_1)(c)=1$ induces a probability of reaching $s_2$ of $0.36$, see the induced MC in Fig.~\ref{fig:induced_dtmc1}. For the randomized strategy $\sched_{\textit{unif}}\in\Sched^\mdp$ with $\sched_{\textit{unif}}(s_0)(\act)=\sched_{\textit{unif}}(s_0)(b)=0.5$ and $\sched_{\textit{unif}}(s_1)(c)=\sched_{\textit{unif}}(s_1)(d)=0.5$, which chooses between all actions uniformly, the specification is also violated: 
%
The probability of reaching $s_2$ is $0.25$, hence $\sched_{2}\not\models\phi$. 
However, for the deterministic strategy $\sched_{\textit{safe}}\in\Sched^\mdp$ with $\sched_{\textit{safe}}(s_0)(b)=1$ and $sched_{\textit{safe}}(s_1)(d)=1$ the probability is $0.16$, thus $\sched_{\textit{safe}}\models\phi$. Note that $\sched_{\textit{safe}}$ minimizes the probability of reaching $s_2$ while $\sched_1$ maximizes this probability.
\end{example}

\section{Synthesizing shared control protocols}\label{sec:synthesisproblem}
\noindent In this section we describe our formal approach to synthesize a shared control protocol in presence of randomization. We start by formalizing the concepts of \emph{blending} and \emph{strategy perturbation}. Afterwards we formulate the general problem and show that the solution to  the synthesis problem is correct.
%

\begin{example}\label{ex:wheelchair}
\newcommand{\gridScale}{1} 

\newcommand\mygrid[1]{
    \draw[black,line width=1pt] (0,0) grid[step=1] (#1,#1);
    \draw[black,line width=4pt] (0,0) rectangle (#1,#1);
}

\newcommand{\fillGridAt}[3]{
	\node [xshift=.5*\gridScale cm,yshift=.5*\gridScale cm] at (#1,#2){#3};	
}

\renewcommand{\gridScale}{1}
\newcommand{\robotScale}{0.03}
\newcommand{\flagScale}{0.1}
\newcommand{\broomScale}{0.05}

\begin{figure}[t]
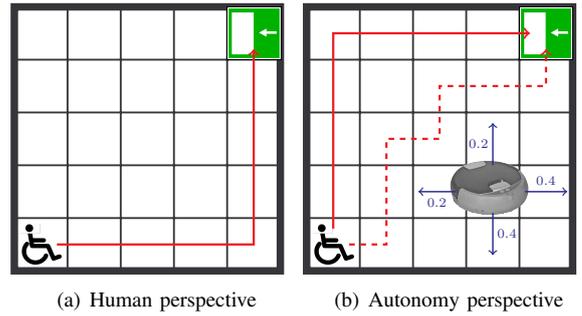

	\centering
	\subfigure[Human perspective]
	{%
		\scalebox{0.7}{\input{pics/wheelchair_human}}
		\label{fig:gridworld_human}
	}%
	\subfigure[Autonomy perspective]
	{%
		\scalebox{0.7}{\input{pics/wheelchair_full}}		
		\label{fig:gridworld_full}
	}%
	\caption{A wheelchair in a shared control setting.}
	\label{fig:gridworld}
\end{figure}
Consider Fig.~\ref{fig:gridworld}, where a room to navigate in is abstracted into a grid. We will use this as our ongoing example. A wheelchair as in~\cite{wheelchair-demillan} is to be steered from the lower left corner of the grid to the exit on the upper right corner of the grid. There is also an autonomous robotic vacuum cleaner moving around the room; the goal is for the wheelchair to reach the exit without crashing into the vacuum cleaner. We now assume that the vacuum cleaner moves according to probabilities that are fixed according to evidence gathered beforehand; these probabilities are unknown or incomprehensible to the human user. To improve the safety of the wheelchair, it is equipped with an autonomy protocol that is to improve decisions of the human or even overwrite them in case of safety hazards. For the design of the autonomy protocol, the evidence data about the cleaner is present.

Now an obvious strategy to move for the wheelchair, not taking into account the vacuum cleaner, is depicted by the red solid line in Fig.~\ref{fig:gridworld_human}. As indicated in Fig.~\ref{fig:gridworld_full}, the strategy proposed by the human is \emph{unsafe} because there is a high probability to collide with the obstacle. The autonomy protocol computes a safe strategy, indicated by the solid line in Fig.~\ref{fig:gridworld_full}. As this strategy deviates highly from the human strategy, the dashed line indicates a still safe enough alternative which is a compromise or---in our terminology---a blending between the two strategies.
\end{example}
We assume in the following that possible \emph{behaviors of the robot} inside the environment are modeled by MDP $\mdp_r=(S,\sinit,\Act,\probmdp)$. The \emph{human strategy} is given as randomized strategy $\sched_h$ for $\mdp_r$. We explain how to obtain this strategy in Section~\ref{sec:simulation}. \emph{Specifications} are $\varphi_1,\ldots,\varphi_n$ being either \emph{safety properties} $\reachProplT$ or \emph{performance properties} $\expRewProp{\kappa}{T}$.
\subsection{Strategy blending} \noindent Given two strategies, they are to be \emph{blended} into a new strategy favoring decisions of one or the other in each state of the MDP. In our setting, the human strategy $\sigma_h\in\Sched^{\mdp_r}$ is blended with the autonomous strategy $\sigma_a\in\Sched^{\mdp_r}$ by means of an arbitrary \emph{blending function}. In~\cite{dragan-et-al-policy-blending} it is argued that blending intuitively reflects the \emph{confidence} in how good the autonomy protocol is able to assist with respect to the human user's goals. In addition, factors probably unknown or incomprehensible for the human such as safety or performance optimization also should be reflected by such a function. 

Put differently, possible actions of the user should be assigned \emph{low confidence} by the blending function, if he cannot be trusted to make the right decisions.
For instance, recall Example~\ref{ex:wheelchair}.
At cells of the grid where with a very high probability the wheelchair might collide with the vacuum cleaner, it makes sense to assign a high confidence in the autonomy protocol's decisions because not all safety-relevant information is present for the human. 

In order to enable formal reasoning together with such a function we instantiate the blending with a \emph{state-dependent} function which at each state of an MDP weighs the confidence in both the human's and the autonomy's decisions. A more fine-grained instantiation might incorporate not only the current state of the MDP but also the strategies of both human and autonomy or history of a current run of the system. Such a formalism is called \emph{linear blending} and is used in what follows. In~\cite{DBLP:journals/corr/Trautman15}, additional notions of blending are discussed. 

\begin{definition}[Linear blending]\label{def:blending}
Given an MDP $\mdp_r=(S,\sinit,\Act,\probmdp)$, two strategies $\sched_h,\sched_a\in\Sched^{\mdp_r}$, and a \emph{blending function} $\blendFunc\colon S\rightarrow [0,1]$,  \emph{the blended strategy $\sigma_{ha}\in\Sched^{\mdp_r}$} for all states $s\in S$, and actions $\act\in\Act$ is
			\begin{align*}
				\sigma_{ha}(s)(\alpha)=\blendFunc(s)\cdot\sched_h(s)(\alpha) + (1-\blendFunc(s))\cdot\sched_a(s)(\alpha)\ .
			\end{align*}			
\end{definition}
\smallskip
Note that the blended strategy $\sigma_{ha}$ is a well-defined randomized strategy. For each $s\in S$, the value $\blendFunc(s)$ represents the confidence in the human's decisions at this state, \ie, the ``weight'' of $\sigma_h$ at $s$. 

Coming back to Example~\ref{ex:wheelchair}, the critical cells of the grid correspond to certain states of the MDP $\mdp_r$; at these states a very low confidence in the human's decisions should be assigned. For instance at such a state $s\in S$ we might have $b(s)=0.1$ leading to the fact that all randomized choices of the human strategy are scaled down by this factor. Choices of the autonomous strategy are only scaled down by factor $0.9$. The addition of these scaled choices then gives a new strategy highly favoring the autonomy's decisions.

%
%
%
\subsection{Perturbation of strategies}
\noindent As mentioned before, we want to ensure that the blended strategy deviates minimally from the human strategy. To now \emph{measure} such a deviation, we introduce the concept of \emph{perturbation} which was---on a complexity theoretic level---for instance investigated in~\cite{chen-et-al-concur-2014-pertubation}. 
	Here, we introduce an \emph{additive perturbation} for a (randomized) strategy, incrementing or decrementing probabilities of action choices such that a well-defined  distribution over actions is maintained.
\begin{definition}[Strategy perturbation]\label{def:perturbation}
	Given MDP $\mdp$ and strategy $\sched\in\Sched^\mdp$, an (additive) \emph{perturbation} $\delta$ is a function $\delta\colon S\times\Act\rightarrow[-1,1]$ with
	\begin{align*}
		\sum_{\act\in\Act}\delta(s,\act)=0 \text{ for all } s\in S\ . 
	\end{align*}	
	The value $\delta(s,\act)$ is called the \emph{perturbation value} at state $s$ for action $\act$. Overloading the notation, the \emph{perturbed strategy} $\delta(\sched)$ is given by
	\begin{align*}
		\delta(\sched)(s,\act)=\sched(s)(\act)+\delta(s,\act) \text{ for all } s\in S \text{ and } \act\in\Act \ .
	\end{align*}
\end{definition}
%
%
%
%
\subsection{Design of the autonomy protocol} 
\noindent For the \emph{formal problem}, we are given blending function $\blendFunc$, specifications $\varphi_1,\ldots,\varphi_n$, MDP $\mdp_r$, and human strategy $\sigma_h\in\mdp_r$. We assume that $\sigma_{h}$ does not satisfy all of the specifications, \ie, $\sigma_h\not\models\varphi_1,\ldots,\varphi_n$. The \emph{autonomy protocol} provides the \emph{autonomous strategy} $\sched_a\in\Sched^{\mdp_r}$. According to $\blendFunc$, the strategies $\sigma_a$ and $\sigma_h$ are blended into strategy $\sigma_{ha}$, see Definition~\ref{def:blending}, \ie, $\sigma_{ha}(s,\alpha)=\blendFunc(s)\cdot\sched_a(s,\alpha) + (1-\blendFunc(s))\cdot\sched_h(s,\alpha)$. The \emph{shared control synthesis problem} is to design the autonomy protocol such that for the blended strategy $\sigma_{ha}$ it holds $\sigma_{ha}\models\varphi_1,\ldots,\varphi_n$, while \emph{minimally deviating} from $\sigma_h$. The deviation from $\sigma_h$ is captured by finding a perturbation $\delta$ as in Definition~\ref{def:perturbation}, where, \eg, the \emph{infinity norm} of all perturbation values is minimal.
 	
		\noindent Our problem involves the explicit computation of a randomized strategy  and the induced probabilities, which is inherently nonlinear because the corresponding variables need to be multiplied. Therefore, the canonical formulation is given by a nonlinear optimization program (NLP). We first assume that the only specification is a quantitative reachability property $\varphi=\reachProplT$, then we describe how more properties can be included.
		The program has to encompass defining the autonomous strategy $\sched_a$, the perturbation $\delta$ of the human strategy, the blended strategy $\sched_{ha}$, and the probability of reaching the set of target states $T\subseteq S$. 
		
			We introduce the following specific set $\Vars$ of \emph{variables}:
		\begin{itemize}
			\item $\sched_{a}^{s,\alpha}, \sched_{ha}^{s,\alpha}\in[0,1]$ for each $s\in S$ and $\act\in\Act$ define the autonomous strategy $\sched_a$ and the blended strategy $\sched_{ha}$. 
			\item $\delta^{s,\alpha}\in[-1,1]$ for each $s\in S$ and $\act\in\Act$ are the perturbation variables for $\sched_h$ and $\sched_{ha}$.
		\item $p_s\in[0,1]$ for each $s\in S$ are assigned the probability of reaching $T\subseteq S$ from state $s$ under strategy $\sched_{ha}$.
		\end{itemize}
	 	Using these variables, the NLP reads as follows:
		\begin{align}
			\text{minimize } &\quad \max \{|\delta^{s\act}| \mid s\in S,\act\in\Act \} \label{eq:strategyah:min}\\
			\text{subject to}
							 &\quad p_{\sinit}\leq \lambda\label{eq:strategyah:lambda}\\
			\forall s\in T.	 &\quad p_s=1\label{eq:strategyah:targetprob}\\
			\forall s\in S.	&\quad \sum_{\act\in\Act}\sched_{a}^{s,\act}=\sum_{\act\in\Act}\sched_{ha}^{s,\act}=1\label{eq:strategyah:welldefined_sched}\\
			\forall s\in S.\,\forall \act\in\Act. &\quad \sched_{ha}^{s,\act}=\sched_h(s)(\act)+\delta^{s,\act}\label{eq:strategyah:perturbation}\\
			\forall s\in S.	&\quad \sum_{\act\in\Act} \delta^{s,\act}=0\label{eq:strategyah:perturbation_welldefined}\\
			\forall s\in S.\,\forall \act\in\Act. &\quad \sched_{ha}^{s,\act}=\blendFunc(s)\cdot\sched_h(s)(\act) + \left(1-\blendFunc(s)\right)\cdot\sched_{a}^{s,\act}\label{eq:strategyah:blending}\\
			\forall s\in S.	&\quad p_s=\sum_{\act\in\Act}\sigma_{ha}^{s,\act}\cdot\sum_{s'\in S}	\probmdp(s,\act)(s')\cdot p_{s'}\label{eq:strategyah:probcomputation}	
		\end{align}
		
	The NLP works as follows. First, the infinity norm of all perturbation variables is minimized (by minimizing the maximum of all perturbation variables)~\eqref{eq:strategyah:min}. The probability assigned to the initial state $\sinit\in S$ has to be smaller than or equal to $\lambda$ to satisfy $\varphi=\reachProplT$~\eqref{eq:strategyah:lambda}. For all target states $T\subseteq S$, the probability of the corresponding probability variables is assigned one~\eqref{eq:strategyah:targetprob}. Now, to have well-defined strategies $\sched_{a}$ and $\sched_{ha}$, we ensure that the assigned values of the corresponding strategy variables at each state sum up to one~\eqref{eq:strategyah:welldefined_sched}. The perturbation $\delta$ of the human strategy $\sched_h$ resulting in the strategy $\sched_{ha}$ as in Definition~\ref{def:perturbation} is computed using the perturbation variables~\eqref{eq:strategyah:perturbation}; in order for the perturbation to be well-defined, the variables have to sum up to zero at each state~\eqref{eq:strategyah:perturbation_welldefined}.
	The blending of $\sched_a$ and $\sched_{ha}$ with respect to $\blendFunc$ as in Definition~\ref{def:blending} is defined in~\eqref{eq:strategyah:blending}.
	Finally, the probability to reach $T\subseteq S$ from each $s\in S$ is computed in~\eqref{eq:strategyah:probcomputation}, defining a non-linear equation system, where action probabilities, given by the induced strategy $\sched_{ha}$, are multiplied by probability variables for all possible successors. 
		
Note that this nonlinear program is in fact \emph{bilinear} due to multiplying the strategy variables $\sigma_{ha}^{s,\act}$ with the probability variables $p_{s'}$~\eqref{eq:strategyah:probcomputation}. The number of constraints is governed by the number of state and action pairs, \ie, the \emph{size of the problem} is in $\mathcal{O}(|S_r|\cdot |\Act|)$.

An \emph{assignment} of real-valued variables is a function $\ass\colon\Vars\rightarrow\R$; it is \emph{satisfying} for a set of (in)equations, if each one evaluates to \true. A satisfying assignment $\ass^*$ is \emph{minimizing} with respect to objective $o$ if for $\ass^*(o)\in\R$ there is no other assignment $\ass'$ with $\ass'(o)<\ass^*(o)$. Using these notions, we state the correctness of the NLP in~\eqref{eq:strategyah:min} -- \eqref{eq:strategyah:probcomputation}.

\begin{theorem}[Soundness and completeness]\label{theo:correctness}
	The NLP 
	is \emph{sound} in the sense that each minimizing assignment induces a solution to the shared control synthesis problem. It is \emph{complete} in the sense that for each solution to the shared control synthesis there is a minimizing assignment of the NLP.
\end{theorem}\smallskip
Soundness tells that each satisfying assignment of the variables corresponds to strategies $\sched_a$ and $\sched_{ha}$ as well as the perturbation $\delta$ as defined above. Moreover, any optimal solution induces a perturbation minimally deviating from the human strategy $\sched_h$.
%
%
Completeness means that all possible solutions of the shared control synthesis problem can be encoded by this NLP. Unsatisfiability means that no such solution exists; the problem is \emph{infeasible}.

\subsection{Additional specifications}
\noindent We now explain how the NLP can be extended for further specifications. Assume in addition to $\varphi=\reachProplT$, another reachability property $\varphi'=\reachProp{\lambda'}{T'}$ with $T'\neq T$ is given. We add another set of probability variables $p'_s$ for each state $s\in S$;~\eqref{eq:strategyah:lambda} is copied for $p'_{\sinit}$ and $\lambda'$,~\eqref{eq:strategyah:targetprob} is defined for all states $s\in T\cup T'$ and~\eqref{eq:strategyah:probcomputation} is copied for all $p'_s$, thereby computing the probability of reaching $T'$ under $\sched_{ha}$ for all states.

	To handle an \emph{expected cost property} $\expRewProp{\kappa}{G}$ for $G\subseteq S$, we use variables $r_s$ being assigned the expected cost for reaching $G$ for all $s\in S$. We add the following equations:
	\begin{align}
							 &\quad r_{\sinit}\leq \kappa\label{eq:strategyahexp:kappa}\\
			\forall s\in G.	 &\quad r_s=0\label{eq:strategyahexp:targetrew}\\
			\forall s\in S.	&\quad r_s= \sum_{\act\in\Act} \Bigl(\sigma_{ha}^{s,\act} \cdot \rew(s,\act) +    \sum_{s'\in S}	\probmdp (s,\act)(s') \cdot r_{s'}\Bigr)\label{eq:strategyahexp:rewcomputation}	
		\end{align}
		First, the expected cost of reaching $G$ is smaller than or equal to $\kappa$ at $\sinit$~\eqref{eq:strategyahexp:kappa}. Goal state are assigned cost zero~\eqref{eq:strategyahexp:targetrew}, otherwise infinite cost is collected at absorbing states. Finally, the expected cost for all other states is computed by~\eqref{eq:strategyahexp:rewcomputation} where according to the blended strategy $\sched_{ha}$ the cost of each action is added to the expected cost of the successors. 
	An important insight is that if all specifications are expected reward properties, the program is \emph{no longer nonlinear} but a linear program (LP), as there is no multiplication of variables.
\setlength{\tabcolsep}{4pt}
\begin{table}[t]\centering\caption{Example Results}
		\scalebox{0.8}{
		\begin{tabular}{@{}ccccccccccc@{}}
		\toprule
  		& $b_i$ & $\sched_a(a)$ & $\sched_a(b)$ & $\sched_a(c)$ & $\sched_a(d)$ & $\sched_{ha}(a)$ & $\sched_{ha}(b)$ & $\sched_{ha}(c)$ & $\sched_{ha}(d)$ & $\pr_{ha}$\\
  		\midrule
  		$i=1$ & $0.5$  & $0.08$ & $0.92$ & $0.08$ & $0.92$ & $0.29$ & $0.71$ & $0.29$ & $0.71$ & $0.209$\\
  		$i=2$ & $0.1$  & $0.27$ & $0.73$ & $0.27$ & $0.73$ & $0.29$ & $0.71$ & $0.29$ & $0.71$ & $0.209$\\
  		$i=3$ & $0$ & $0.29$ & $0.71$ & $0.29$ & $0.71$ & $0.29$ & $0.71$ & $0.29$ & $0.71$ & $0.209$\\
		\bottomrule\end{tabular}}
		\label{tab:example}	
	\end{table}
\subsection{Generalized blending} 
If the problem is not feasible for the given blending function, optionally the autonomy protocol can try to compute a new function $\blendFunc\colon S\rightarrow [0,1]$ for which the altered problem is feasible. We call this procedure \emph{generalized blending}. The idea is that computing this  function gives the designer of the protocol insight on where more confidence needs to be placed into the autonomy or, vice versa, where the human cannot be trusted to satisfy the given specifications.

Computing this new function is achieved by nearly the same NLP as for a fixed blending function while adding variables $\blendFunc^s$ for each state $s\in S$, defining the new blending function by $\blendFunc(s)=\blendFunc^{s}$. We substitute Equation~\ref{eq:strategyah:blending} by 
	\begin{align}
		\forall s\in S.\,\forall \act\in\Act. &\quad \sched_{ha}^{s,\act}=\blendFunc^s\cdot\sched_h(s)(\act) + \left(1-\blendFunc^s\right)\cdot\sched_{a}^{s,\act}\ .\label{eq:strategyah:generalblending}
	\end{align}
	A satisfying assignment for the resulting nonlinear program induces a suitable blending function $\blendFunc\colon S\rightarrow[0,1]$ in addition to the strategies. If this problem is also infeasible, there is no strategy that satisfies the given specifications for MDP $\mdp_r$.
\begin{corollary}
If there is no solution for the NLP given by Equations~\ref{eq:strategyah:min} --~\ref{eq:strategyah:generalblending}, there is no strategy $\sched\in\Sched^{\mdp_r}$ such that $\sched\models\varphi_1,\ldots,\varphi_n$.
\end{corollary}
As there are no restrictions on the blending function, this corollary trivially holds: Consider for instance $\blendFunc$ with $\blendFunc(s)=0$ for each $s\in S$. This function disregards the human strategy which may be perturbed to each other strategy $\sched_a=\sched_{ha}$.
\begin{example}\label{ex:nlpex}
Reconsider the MDP $\mdp$ from Example~\ref{ex:simple_mdp} with  specification $\varphi=\reachProp{0.21}{\{s_2\}}$ and the randomized strategy $\sched_{\textit{unif}}$ which takes each action uniformly distributed. As we saw, $\sched_{\textit{unif}}\not\models\varphi$. We choose this strategy as the human strategy $\sched_h=\sched_{\textit{unif}}$ and $\mdp_r=\mdp$ as the robot MDP. For a blending function $b_h$ putting high confidence in the human, \eg, if $b_h(s)\geq 0.6$ for all $s\in S$, the problem is infeasible. 

In Table~\ref{tab:example} we display results putting medium ($\blendFunc_1$), low ($\blendFunc_2$), or no confidence ($\blendFunc_3$) in the human at $s_0$ and $s_1$. We list the assignments for the resulting strategies $\sched_a$ and $\sched_{ha}$ as well as the probability $\pr_{ha}=\pr_{s_0}^{\mdp_r^{\sched_{ah}}}(\finally T)$ to reach $s_2$ under the blended strategy $\sched_{ha}$. The results were obtained using the NLP solver \tool{IPOPT}~\cite{ipopt}.

We observe that for decreasing confidence in the human decisions, the autonomous strategy has higher probabilities for actions $a$ and $c$ which are the ``bad'' actions here. That means that---if there is a higher confidence in the autonomy---solutions farer away from the optimum are good enough. The maximal deviation from the human strategy is $0.21$. \emph{Generalized blending} with maximizing over the confidence in the human's decisions at all states $s\in S$ yields $b_h(s)=0.582$, \ie, we compute the highest possible confidence in the human's decisions where the problem is still feasible under the given human strategy.
\end{example}

%


\section{Computationally Tractable Approach}\label{sec:solution}
\noindent The nonlinear programming approach presented in the previous section gives a rigorous method to solve the shared control synthesis problem and serves as mathematically concise definition of the problem. However, NLPs are known to have severe restrictions in terms of scalability and suffer from numerical instabilities. 
%
The crucial point to an efficient solution is circumventing the expensive computation of optimal randomized strategies and reducing the number of variables. We propose a heuristic solution which enables to use linear programming (LP) while ensuring soundness.


We utilize a technique referred to as \emph{model repair}. Intuitively, an erroneous model is changed such that it satisfies certain specifications. In particular, given a Markov chain $\dtmc$ and a specification $\varphi$ that is violated by $\dtmc$, a repair of $\dtmc$ is an automated method that transforms it to new MC $\dtmc'$ such that $\varphi$ is satisfied for $\dtmc'$. Transforming refers to changing probabilities or cost while regarding certain side constraints such as keeping the original graph structure.

In~\cite{bartocci2011model}, the first approach to automatically repair an MC model was presented as an NLP. Simulation-based algorithms were investigated in~\cite{chen2013model}. A heuristic but very scalable technique called \emph{local repair} was proposed in~\cite{pathak-et-al-nfm-2015}. This approach greedily changes the probabilities or cost of the original MC until a property is satisfied. An upper bound $\delta_r$ on changes of probabilities or cost can be specified; correctness and completeness can be given in the sense that if a repair with respect to $\delta_r$ exists, it will be obtained. 
%
%
%

Take now the MC $\dtmc_r^{\sched_h}$ which is induced by the robot MDP $\mdp_r$ and the human strategy $\sched_h$. We perform model repair such that the repaired MC $\dtmc'=(S,\sinit,P')$ satisfies the specifications $\varphi_1,\ldots,\varphi_n$. The question is now, how from the repaired MC $\dtmc'$, the strategy $\sched'\in\Sched^{\mdp_r}$ can be extracted. More precisely, we need $\sched'$ inducing exactly $\dtmc'$, \ie, $\dtmc_r^{\sigma'}=\dtmc'$, when applied to MDP $\mdp_r$. 

First, we need to make sure that the repaired MC is \emph{consistent} with the original MDP such that a strategy $\sched'$ with $\dtmc_r^{\sigma'}=\dtmc'$ actually exists. Therefore, we define the maximal and minimal possible transition probabilities $P_{\max}$ and $P_{\min}$  that can occur in any induced MC of MDP $\mdp_r$:
	\begin{align}
		P_{\max}(s,s') = \max\{\probmdp_r(s,\act)(s')\mid\act\in\Act\}  
	\end{align}
for all $s\in S$; $P_{\min}$ is defined analogously. Now, the repair is performed such that in the resulting MC $\dtmc'=(S,\sinit,P')$ for all $s,s'\in S$ it holds that 
\begin{align}
	P_{\min}(s,s') \leq P(s,s') \leq P_{\max}(s,s') \ .
\end{align}
While obtaining $\dtmc'$, model checking needs to be performed intermediately to check if the specifications are satisfied; once they are, the algorithm terminates.
 In fact, for each state $s\in S$, the probability of satisfaction is computed. We assign variables $\mathit{mc}_s$ for all $s\in S$ with exactly this probability:
\begin{align}
	\mathit{mc}_s=\pr(s\models\varphi_1,\ldots,\varphi_n)\ .	
\end{align}
Now recall the NLP from the previous section, in particular Equation~\ref{eq:strategyah:probcomputation} which is the only nonlinear equation of the program. We replace each variable $p_s$ by the concrete model checking result $\mathit{mc}_s$ for each $s\in S$:
 \begin{align}
	 \mathit{mc}_s=\sum_{\act\in\Act}\sigma_{ha}^{s,\act}\cdot\sum_{s'\in S}	\probmdp(s,\act)(s')\cdot \mathit{mc}_{s'}\ .\label{eq:strategyah:probcomputation_lp} 	
 \end{align}
As~\eqref{eq:strategyah:probcomputation_lp} is affine in the variables $\sched_{ah}$, the program resulting from replacing \eqref{eq:strategyah:probcomputation} by \eqref{eq:strategyah:probcomputation_lp} is a linear program (LP). Moreover, \eqref{eq:strategyah:lambda} and \eqref{eq:strategyah:targetprob} can be removed, reducing the number of constraints and variables. The LP gives a feasible solution to the shared control synthesis problem.
\begin{lemma}[Correctness]\label{cor:modelrepair}
	The LP is sound in the sense that each minimizing assignment induces a solution to the shared control problem.
\end{lemma}
The correctness is given by construction, as the specifications are satisfied for the blended strategy which is derived from the repaired MC.	
However, the minimal deviation from the human strategy as in Equation~\ref{eq:strategyah:min} is dependent on the previous computation of probabilities for the blended strategy. Therefore, we actually compute an \emph{upper bound} on the optimal solution. Let $\delta^*$ be the minimal deviation possible for any given problem and $\delta$ be the minimal deviation obtained by the LP resulting from replacing~\eqref{eq:strategyah:probcomputation} by~\eqref{eq:strategyah:probcomputation_lp}. Let $\|\delta\|_{\infty}$ and $\|\delta^*\|_{\infty}$ denote the infinity norms of both perturbations.
\begin{corollary}
	For the perturbations $\delta$ and $\delta^*$ of $\sched_h$ it holds that
			$\|\delta^*\|_{\infty} \leq \|\delta\|_{\infty}$.
\end{corollary}
As we mentioned before, the local repair method can employ a bound $\delta_r$ on the maximal change of probabilities or cost in the model. If a repair exists for a given $\delta_r$, the resulting deviation $\delta$ is then bounded by this $\delta_r$.

\section{Case study and experiments}\label{sec:simulation}
\noindent Defining a formal synthesis approach to the shared control scenario requires a precomputed estimation of a human user's intentions. As explained in the previous chapter, we account for inherent uncertainties by using a \emph{randomized strategy} over possible actions to take. We discuss how such strategies may be obtained and report on benchmark results.

\subsection{Experimental setting}
\noindent Our setting is the wheelchair scenario from Example~\ref{ex:wheelchair} inside an interactive \tool{Python}  environment. The size of the grid is variable and an arbitrary number of stationary and randomly moving obstacles (the vacuum cleaner) can be defined. An agent (the wheelchair) is moved according to predefined (randomized) strategies or interactively by a human user.

From this scenario, an MDP with states corresponding to the position of the agent and the obstacles is generated. Actions induce position changes of the agent. The safety specification ensures that the agent reaches a target cell without crashing into an obstacle with a certain high probability $\lambda\in[0,1]$, formally $\pctlProb_{\geq \lambda}(\neg \texttt{crash } \pctlUntil \texttt{ target})$. We use the probabilistic model checker \tool{PRISM}~\cite{KNP11} for verification, in form of either a worst--case analysis for each possible strategy or concretely for a specific strategy. The whole toolchain integrates the simulation environment with the approaches described in the previous sections. We use the NLP solver \tool{IPOPT}~\cite{ipopt} and the LP solver \tool{Gurobi}~\cite{gurobi}. To perform model repair for strategies, see Section~\ref{sec:solution}, we implemented the greedy method from~\cite{pathak-et-al-nfm-2015} into our framework augmented by side constraints ensuring well-defined strategies.

\subsection{Data collection}
\noindent We ask five participants to perform tests in the  environment with the goal to move the agent to a target cell while never being in same cell as the moving obstacle. From the data obtained from each participant, an individual randomized human strategy $\sched_h$ for this participant can be obtained via Maximum Entropy Inverse Reinforcement Learning (MEIRL)~\cite{ziebart2008maximum}. Inverse reinforcement learning has---for instance---also been used in~\cite{javdani2015shared} to collect data about human behavior in a shared control scenario (though without any formal guarantees) or in~\cite{rothkopf2013modular} to distinguish human intents with respect to different tasks.
 In our setting, each \emph{sample} is one particular command of the participant, while we have to assume that command is actually made with the intent to satisfy the specification of safely reaching a target cell. For the resulting strategy, the probability of a possible deviation from the actual intend can be bounded with respect to the number of samples using Hoeffding's inequality, see~\cite{ziebart2010modeling} for details. On the other hand, we can determine the number of samples needed to get a reasonable approximation of typical behavior. 
 
The concrete probabilities of possible deviation depend on $\mathcal{O}(\exp(-n\epsilon))$, where $n$ is the number of samples and $\epsilon$ is the desired upper bound on the deviation between the true probability of satisfying the specification and the average obtained by the sampled data.
Here, in order to ensure an upper bound $\epsilon = 0.1$ with probability $0.99$, the required amount of samples is $265$.
\begin{figure}[t]
	\centering\scalebox{0.7}{
\begin{tikzpicture}
\tikzstyle{outer}= [draw, text centered, shape=rectangle, text width=1.8cm]
\tikzstyle{inner}=[draw, text centered, shape=rectangle, rounded corners, text width=2cm, minimum height=1.1cm, inner sep=5pt]			

\node[inner, very thick] (learning) {Process data via MEIRL};

\node[inner, right=1.6cm of learning, very thick, text width=2.5cm] (synthesis) {Shared control synthesis};

\node[inner, below=1.6cm of synthesis, very thick] (simulation) {Simulation environment};

\node[inner, right=1.6cm of synthesis, very thick] (blending) {Blending function};

%
%

\draw[thick,-latex'] (learning) -- node[below, text centered, text width=1.2cm] {human strategy} (synthesis);
\draw[thick,-latex'] (synthesis) -- node[auto, text width=1.2cm, text centered] {autonomous strategy} (simulation);
\draw[thick,-latex'] (blending) -- (synthesis);
\draw[thick,-latex'] (blending) |- (simulation);
\draw[thick,-latex'] (simulation) -| node[above, near start, text width=1.2cm, text centered] {sample data} (learning);

%
%
%
%
%
%
\node (box) [draw=white, fit = (current bounding box), inner sep=0.2cm] {};

\end{tikzpicture}}
	\caption{Experimental setting for the shared control simulation.}	
	\label{fig:data}
\end{figure}
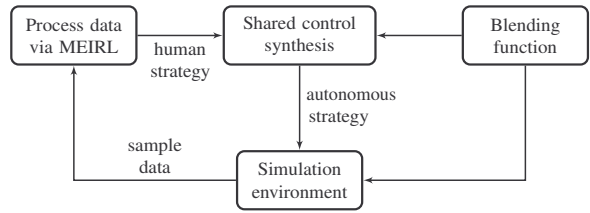
\subsection{Experiments}
\noindent The work flow of the experiments is depicted in Figure~\ref{fig:data}. First off, we discuss sample data for one particular participant using a $8\times 8$ grid with one moving obstacle inducing an MDP of $2304$ states. 
In the synthesis, we employ the model repair procedure as explained in Section~\ref{sec:solution} because the approach based on NLP is only feasible for very small examples. We design the blending function as follows: At states where the human strategy induces a high probability of crashing, we put low confidence in the human and vice versa. Using this function, the autonomous strategy $\sched_a$ is created and passed (together with the function) back to the environment. Note that the blended strategy $\sched_{ah}$ is ensured to satisfy the specification, see Lemma~\ref{cor:modelrepair}. Now, we let the same participant as before do test runs, but this time we blend the human commands with the (randomized) commands of the autonomous strategy $\sched_a$. Then the actual action of the agent is determined stochastically. 
We obtain the following results. Our safety specification is instantiated with $\lambda=0.7$, ensuring that the target is safely reached with at least probability $0.7$. The human strategy $\sched_{h}$ has probability $0.546$, violating the specification. With the aforementioned blending function we compute $\sched_a$ which induces  probability $0.906$. Blending these two strategies into $\sched_{ah}$ yields a probability of $0.747$. When testing the synthesized autonomy protocol for the individual participant, we observe that his choices are mostly corrected if intentionally bad decisions are made. Also, simulating the blended strategy leeds to the expected result that the agent does not crash in roughly $70\%$ of the cases.

\begin{figure}
	\subfigure[strategy $\sched_h$]{
		\includegraphics[scale=0.4]{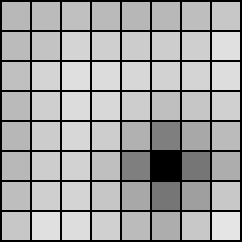}
	}
		\subfigure[strategy  $\sched_{ah}$]{
		\includegraphics[scale=0.4]{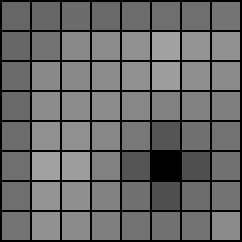}
	}
	\subfigure[strategy $\sched_h$]{
		\includegraphics[scale=0.4]{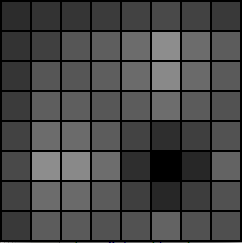}
	}
	\caption{Graphical representation of the obtained human, blended, and autonomous strategy in the grid.}
	\label{fig:heatmap}
\end{figure}

To make the behavior of the strategies more accessible, consider Figure~\ref{fig:heatmap}. For each $\sched_a$, $\sched_{ah}$, and $\sched_h$ we indicate for each cell of the grid the worst-case probability to safely reach the target. This probability depends on the current position of the obstacle, which is again probabilistic. The darker the color, the higher the probability; thereby black indicates a probability of $1$ to reach the target. We observe that the human's decisions are rather risky even near the target, while for the blended strategy---once the agent is near the target---there is a very high probability of reaching it safely. This representation also shows that with our approach the blended strategy improves the human strategy while not changing it too much. Specifically, the maximal deviation from the human strategy is $0.27$, which is the result of the infinity norm as in Equation~\ref{eq:strategyah:min}.

To finally assess the  scalability of our approach, consider Table~\ref{tab:scalability}. We generated  MDPs for several grid sizes, number of obstacles, and human strategies. We list the number of reachable MDP states (states) and the number of transitions (trans.). We report on the time the synthesis process took  (synth.), which is basically the time of solving the LP, and the total time including model checking times using \tool{PRISM} (total) measured in seconds. To give an indication on the quality of the synthesis, we list the deviation from the human strategy ($\delta_\infty$). A memory out is indicated by ``--MO--''. All experiments were conducted on a 2.3GHz machine with 8GB of RAM. 
Note that MDPs resulting from grid structures are very strongly connected, resulting in a large number of transitions. Thus, the encoding in the \tool{PRISM}-language~\cite{KNP11} is very large, rendering it a very hard problem. We observe that while the procedure is very efficient for models having a few thousand states and hundreds of thousands of transitions, its scalability is ultimately limited due to memory issues. In the future, we will utilize efficient symbolic data structures internal to \tool{PRISM}. Moreover, we observe that for larger benchmarks the computation time is governed by the solving time of $\tool{Gurobi}$.
\begin{table}[t]
\centering
\caption{Scalability results.}
\label{tab:scalability}
\scalebox{1}{\begin{tabular}{@{}crrrrrrr@{}}
\toprule
grid          & obst. & states     & trans.      & synth. &  total & $\delta_{\infty}$ \\
\midrule
$8\times 8$   & $1$   &  $2.304$    & $36.864$    & $6.30$  & $14.12$ & $0.15$                   \\
$8\times 8$   & $2$   &  $82.944$   & $5.308.416$ & --MO--        & --MO--  & --MO--                 \\
$10\times 10$ & $1$   &  $3.600$    & $57.600$    & $12.29$ & $23.80$ & $0.24$			\\ 
$12\times 12$ & $1$   &  $14.400$   & $230.400$   & $157.94$& $250.78$ & $0.33$                   \\        
\bottomrule\end{tabular}}
\end{table}

\section{Conclusion}\label{sec:conclusion}
\noindent We introduced a formal approach to synthesize autonomy protocols in a shared control setting with guarantees on quantitative safety and performance specifications. The practical usability of our approach was shown by means of data-based experiments. Future work will concern experiments in robotic scenarios and further improvement of the scalability.

\bibliographystyle{plain}
\bibliography{literature}

\end{document}